\PassOptionsToPackage{unicode}{hyperref}
\PassOptionsToPackage{hyphens}{url}
\documentclass[10pt,twocolumn,letterpaper]{article}
\usepackage[pagenumbers]{cvpr}
\makeatletter
\newcommand{\setfigcurrentlabel}{\edef\@currentlabel{\thefigure}}
\newcommand{\settabcurrentlabel}{\edef\@currentlabel{\thetable}}
\makeatother

\newcommand{\mysection}[1]{\vspace{2pt}\noindent{\tenbf #1.}}

\usepackage{iftex}
\ifPDFTeX
  \usepackage[T1]{fontenc}
  \usepackage[utf8]{inputenc}
  \usepackage{textcomp} 
\else 
  \usepackage{unicode-math} 
  \defaultfontfeatures{Scale=MatchLowercase}
  \defaultfontfeatures[\rmfamily]{Ligatures=TeX,Scale=1}
\fi
\ifPDFTeX\else
\fi
\IfFileExists{upquote.sty}{\usepackage{upquote}}{}
\IfFileExists{microtype.sty}{
  \usepackage[]{microtype}
  \UseMicrotypeSet[protrusion]{basicmath} 
}{}
\makeatletter
\@ifundefined{KOMAClassName}{
  \IfFileExists{parskip.sty}{%
    \usepackage{parskip}
  }{
    \setlength{\parindent}{0pt}
    \setlength{\parskip}{6pt plus 2pt minus 1pt}}
}{
  \KOMAoptions{parskip=half}}
\makeatother
\usepackage{array}
\usepackage{calc} 
\makeatletter\@ifundefined{iflatexml}{\newif\iflatexml}{}\makeatother
\iflatexml
\def\legendjustify#1#2#3\par{\caption{#3}}
\makeatletter
\def\fnum@figure{\textbf{\figurename~\thefigure.}}
\def\fnum@table{\textbf{\tablename~\thetable.}}
\def\format@title@figure#1{\lx@tag[][\ ]{\lx@fnum@@{figure}}#1}
\def\format@title@table#1{\lx@tag[][\ ]{\lx@fnum@@{table}}#1}
\makeatother
\else
\def\legendjustify#1#2#3\par{#1{#2}#3\par}
\fi
\iflatexml\renewcommand{\mysection}[1]{\vspace{2pt}\noindent{\normalsize\textbf{#1.}}}\fi

\IfFileExists{footnotehyper.sty}{\usepackage{footnotehyper}}{\usepackage{footnote}}

\makeatletter
\newsavebox\pandoc@box
\newcommand*\pandocbounded[1]{
  \sbox\pandoc@box{#1}%
  \Gscale@div\@tempa{\textheight}{\dimexpr\ht\pandoc@box+\dp\pandoc@box\relax}%
  \Gscale@div\@tempb{\linewidth}{\wd\pandoc@box}%
  \ifdim\@tempb\p@<\@tempa\p@\let\@tempa\@tempb\fi
  \ifdim\@tempa\p@<\p@\scalebox{\@tempa}{\usebox\pandoc@box}%
  \else\usebox{\pandoc@box}%
  \fi%
}
\def\fps@figure{htbp}
\makeatother
\ifLuaTeX
  \usepackage{luacolor}
  \usepackage[soul]{lua-ul}
\else
  \usepackage{soul}
\fi
\usepackage{bookmark}
\IfFileExists{xurl.sty}{\usepackage{xurl}}{} 
\definecolor{cvprblue}{rgb}{0.21,0.49,0.74}
\hypersetup{
  colorlinks=true,
  allcolors=cvprblue,
  pdfcreator={LaTeX via pandoc}}

\title{ByteTraX: Enhancing the ByteTrack Architecture with Optimised
Thresholding}
\iflatexml
\author{Thomas A. O'Shea-Wheller\textsuperscript{1}\\[1.0em]
\textsuperscript{1}University of Exeter}
\date{{\small Corresponding author e-mail: \url{t.a.oshea-wheller@exeter.ac.uk}}}
\else
\author{Thomas A. O'Shea-Wheller\textsuperscript{1}\thanks{Corresponding author e-mail: {\small\urlstyle{tt}\url{t.a.oshea-wheller@exeter.ac.uk}}}\\[1.0em]
\textsuperscript{1}University of Exeter}
\fi

\begin{document}

\maketitle

\begin{abstract}

The ByteTrack algorithm is a widely used and computationally efficient
multi-object tracking architecture. Its core innovation lies in the
combination of lenient bounding box associations with tracklet
similarity matching to robustly deal with object occlusions. However,
this strategy is nevertheless vulnerable to erroneous track
reclassification and identity switching, as detection confidence scores
dictate association priority. To address this, I present a simple
enhancement of the ByteTrack architecture---named ByteTraX---that
optimises track continuity via a single unified matching threshold,
while penalising identity switches through stringent track initiation
criteria. This approach achieves consistently improved performance
across a range of diverse benchmarks including GMOT-40, LC-MOT,
SportsMOT, TeamTrack, DAMUNT, and DeepSea-MOT, while simultaneously
increasing processing speed by \textgreater10\%. Specifically, results
demonstrate a \textgreater40\% reduction in identity switches,
accompanied by mean increases in HOTA of 3.6, IDF1 of 5.6, and FPS of
6.3. As such, adoption of the ByteTraX algorithm has the potential to
substantially enhance tracking performance over the ByteTrack baseline,
while retaining the efficiency needed for real-time deployment. To
facilitate usage, I provide the source code, integration functionality
for the YOLO family of object detection models, and deployment
instructions via an open source repository.

\end{abstract}

\begin{figure*}[t]
{\centering\includegraphics[height=3.44161in]{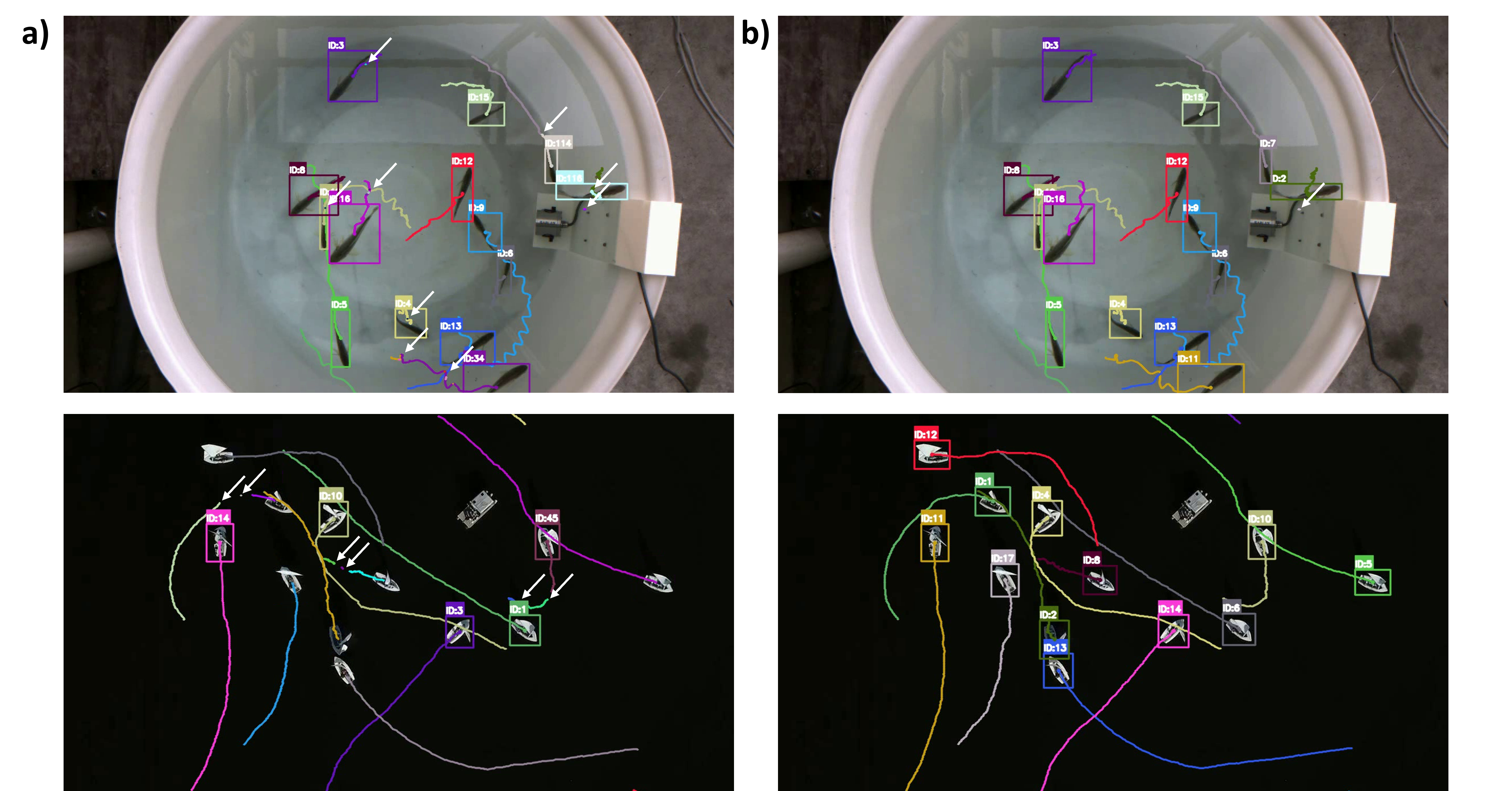}\par}

\vspace{10pt}
{\legendjustify\textbf{Figure 1.} Comparison of tracking ID losses across example video
sequences for (a) ByteTrack, and (b) ByteTraX. The top two panels
display a sequence from LC-MOT, a closed system dataset with fixed track
counts; while the bottom two panels are from GMOT-40, an open system
benchmark in which tracks can enter and leave the frame. Lines indicate
tracked trajectories, coloured by ID, and white arrows denote instances
of ID loss. ID losses occur when bounding box detection scores fall
below a tracking algorithm's matching or similarity threshold, which can
lead to the rapid accumulation of identity errors over time.\par}
\setfigcurrentlabel
\label{fig:1}
\end{figure*}\section{Introduction}

Multi-object tracking algorithms have seen substantial growth in recent
years, with approaches broadly being divided into traditional two-stage
tracking-by-detection methods~\cite{Guan2025,Wojke2017,Cao2023}, and emerging
end-to-end tracking-by-propagation techniques~\cite{Segu2024,Zeng2022,Kim2025}.
While the latter continue to advance the state-of-the-art in absolute
tracking accuracy, algorithms belonging to the former category remain
the most broadly adopted---ostensibly due to their reduced computational
requirements and ease of integration~\cite{Adzemovic2025}. Indeed, when
considering the application of object tracking pipelines more broadly,
accessibility and modularity serve in large part to dictate the
popularity and uptake of a given framework~\cite{Adzemovic2025,Zhang2020}. Consequently,
there is substantial merit in optimising existing
tracking-by-detection algorithms to improve baseline performance, as
this has the potential to yield immediate benefits across established
deployment pipelines.

\begin{figure*}[t]
{\centering\includegraphics[height=2.30688in]{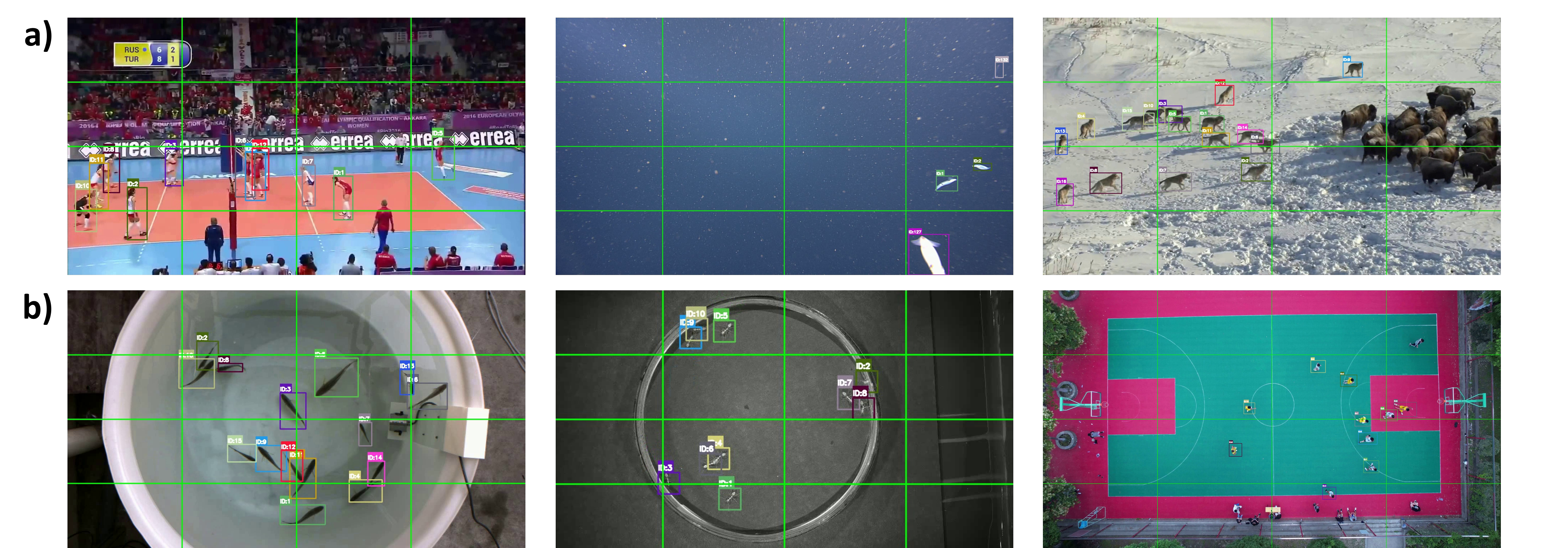}\par}

\vspace{10pt}
{\legendjustify\textbf{Figure 2.} Example video sequences from the (a) open, and (b)
closed system benchmarks used in performance evaluations. The top three
panels detail sequences from SportsMOT, DeepSea-MOT, and GMOT-40; while
the bottom three display sequences from LC-MOT, DAMUNT, and TeamTrack.
Bounding boxes indicate ground truth tracks, coloured by ID, and green
lines denote the crossing counter regions utilised in the calculation of
line crossing accuracy.\par}
\setfigcurrentlabel
\label{fig:2}
\end{figure*}Among current approaches, the ByteTrack~\cite{Zhang2022} algorithm
constitutes one of the most widely utilised architectures, with a
modular design and strong integration into the YOLO~\cite{Redmon2016}
family of deep learning models. Its utility as a tracking solution stems
from its computational efficiency and enhanced occlusion tolerance when
compared to earlier approaches. Specifically, this is achieved by
considering all bounding boxes when associating tracklets---rather than
only those with high confidence scores~\cite{Zhang2022,Zhang2023}. As such,
low confidence detections associated with partially occluded objects are
not erroneously discarded, and thus track continuity is preserved. The
elegance of this approach lies in the understanding that lower
confidence detections may not always represent errors if they adhere to
the algorithm's motion model, and instead may reflect true variance in
an object's appearance.

Despite its versatility, ByteTrack nevertheless exhibits several
limitations in its default configuration. Chief among these are the
propensity for erroneous track reclassification---the assignation of a
new ID to the same object following a brief tracklet loss---and identity
switching---the transposition of IDs between proximate objects (Fig.~\hyperref[fig:1]{1a}). These issues serve to impede long-term track identity preservation,
and rapidly elevate error rates as tracklet length and motion complexity
increase. Interestingly, both are caused by the mistaken elimination or
supersedure of low confidence detections during thresholding, as object
appearances frequently shift during motion. As noted previously, such
low confidence detections sometimes indicate true object positions, and
thus filtering them out can lead to irreversible tracking
errors~\cite{Zhang2022}. The salient aspect of this, however, is that
the tracking algorithm itself has no prior indication of which
confidence distributions are likely to represent true or false
detections for a given model, as both are highly dependent upon upstream
training data and deployment context~\cite{Guo2017}. Consequently,
while low confidence scores generally correlate with the likelihood of
false detections, there is often substantial distributional overlap
between these and true positives in all but the most tightly-fit
models~\cite{ValdenegroToro2021,Grabinski2022}. This fundamentally undermines the
assumption that confidence thresholds alone can reliably discriminate
between true and false detections, thus indicating the need for additional
tracking cues.

With this in mind, there is merit in deferring to the motion and spatial
prediction elements of tracking-by-detection algorithms, as these are
generally model-agnostic and hence less contextually
dependent~\cite{Bashar2026}. In the case of ByteTrack specifically,
this can be achieved by ensuring that confidence-based matching
thresholds are as permissive as possible, hence preserving all candidate
detections that adhere to the motion model. This serves to limit the
potential for erroneous tracklet breaks, and thus associated downstream
identity reassignations. As such, it might be expected that minimising
the thresholds for track association and similarity, while penalising
the generation of new track IDs, could together substantially enhance
baseline tracking performance (Fig.~\hyperref[fig:1]{1b}). However, the optimal approach to
implementing this strategy is dependent upon the attributes of the
systems in which it is deployed. In open systems---where tracks may
enter and exit the frame---threshold adjustment is likely to be the most
effective approach; yet in closed systems---where track numbers
remain constant---there is the opportunity for heuristic-based
implementation. Specifically, this is because in closed systems, global
constants can be leveraged to robustly infer the likelihood that new
tracklets belong to an existing rather than lost track identity. There is thus
impetus to test both approaches as a means to improve tracking fidelity,
as each can be implemented at little to no additional computational
cost.

To this end, I present ByteTraX, an enhancement of the ByteTrack
architecture that maximises the association of low confidence detections
through optimised thresholding, thus limiting the occurrence of
erroneous identity switches. Crucially, ByteTraX considers all
detections equally rather than sorting them by confidence score, predicts
trajectories with broad uncertainty bounds, and deploys bespoke tracking
strategies across open and closed systems. I compare this approach to
the default ByteTrack algorithm across six standardised
benchmarks---GMOT-40~\cite{Bai2021}, LC-MOT~\cite{Yuan2026},
SportsMOT~\cite{Cui2023}, TeamTrack~\cite{Scott2024},
DAMUNT~\cite{Abeysinghe2023}, and
DeepSea-MOT~\cite{Barnard2025}---representing a combination of open and
closed systems from diverse usage scenarios (Fig.~\hyperref[fig:2]{2}). To ensure
consistency, both algorithms employ the same YOLO26 front-end detection
models, and are assessed with standard HOTA, CLEAR, and ID metrics.
Trackers are further evaluated via a simple line crossing accuracy
approach, selected to reflect an explainable and ubiquitous real-world
use case. Results demonstrate that ByteTraX achieves consistent
performance improvements over ByteTrack across benchmarks, while at the
same time delivering increased speed and efficiency.

\section{Related Work}

\subsection{Tracking-by-detection}

While tracking-by-detection approaches have been reviewed extensively
elsewhere~\cite{Guan2025,Adzemovic2025,Luo2021}, there is merit in outlining the
general paradigm as a primer to this work. At its core,
tracking-by-detection functions by associating object detections from an
upstream detection model across frames to form continuous
tracks~\cite{Adzemovic2025}. This method is computationally efficient, can
utilise detections from a diversity of deep learning models, and is
simple to integrate into real-time tracking
pipelines~\cite{Somers2025}. Consequently, a wide range of
architectures exist that employ tracking-by-detection, generally sharing
the following key steps. First, a motion model is used to predict the
future location of tracks via a Kalman filter~\cite{Wojke2017},
particle filter~\cite{Xia2021}, moving horizon
estimate~\cite{Bhatt2023}, or similar method~\cite{Luo2021}.
Second, intersection-over-union (IoU), appearance cues, or a combination
of the two are used to assign similarity scores to high confidence
detections corresponding to the predicted track
location~\cite{Adzemovic2025}. Third, a greedy or Hungarian matching
algorithm is employed to pair one of these candidate detections with the
track based upon its similarity score~\cite{Adzemovic2025,Luo2021}.
Essentially, this can be reduced to the question of where an object is
likely to be in the next frame, what detections overlap with this
projection, and which of these corresponds to the same object given what
is known about its past state.

Despite the effectiveness of this approach, difficulties emerge when
objects are partially occluded, or otherwise fall below the detection
confidence thresholds set by the tracking algorithm or upstream
detection model. Such cases are non-trivial, as the complexity of
real-world tracking encompasses changes in detection confidence that are
frequently independent of actual object
presence~\cite{Mandel2023,Wang2025,Yang2026}. This consequently creates a bottleneck,
as only detections with sufficiently high confidence scores are
considered in the tracking process, meaning that correct object
detections with lower scores are discarded \emph{a priori}.

\subsection{ByteTrack}

The ByteTrack algorithm and BYTE data association method were developed
specifically to address the aforementioned challenge~\cite{Zhang2022}.
While ByteTrack's general structure shares similarities with other
tracking-by-detection approaches, its central innovation lies in how the
BYTE association algorithm filters candidate
detections~\cite{Zhang2022}. Specifically, utilising a two-stage
detection matching technique, BYTE considers both high and low score
bounding boxes in turn, ensuring that partially occluded detections
associated with the latter are recovered. This is achieved by splitting
detections into high and low confidence groups via a fixed threshold,
matching the high confidence detections with associated tracklets, and
then using the remaining low confidence detections to match any
tracklets not associated in the first step. As a consequence, ByteTrack
is able to utilise detections spanning the full range of confidence
scores present in the data, thus avoiding the fundamental limitations
inherent to techniques that consider only those with high scores.

The effectiveness of this method in dealing with partial occlusions and
varying object appearances has positioned ByteTrack as one of the most
widely-utilised tracking algorithms to date~\cite{Adzemovic2025}. However,
in its default configuration, ByteTrack suffers from a high rate of ID
reassignations across trajectories. This is primarily because detection
confidence is still used to score associations and prioritise track
matches, meaning that track breaks and subsequent ID supersedures are
common when the former fluctuates. Here, I argue that reducing the
contribution of detection confidence to matching decisions represents a
simple and robust solution to the issue. As such, I leverage this
approach to develop a novel evolution of the ByteTrack architecture that
prioritises target identity preservation and track continuity through
optimised matching and association thresholds.

\section{Methods}

\subsection{ByteTraX Architecture}

The ByteTraX architecture incorporates two specific enhancements to
improve tracking fidelity and identity persistence. First, it treats all
detections as equally viable by removing the first and second stage
matching thresholds utilised in ByteTrack~\cite{Zhang2022}, meaning
that predictions are not segregated into high and low score groupings.
This ensures that detections with higher confidence scores do not
supersede those of lower confidence unless supported by the motion
model, thus reducing erroneous ID switches when the two sources of
information conflict. This can be calculated as:

\begin{equation}D_{v} = D > \tau\end{equation}

With \(D_{v}\) denoting valid detections used in track matching,
specified as bounding boxes with a confidence score exceeding the
threshold \(\tau\).

Tracked detections are then passed to a central motion model that
predicts their likely coordinates in the next frame, allowing for
propagation of trajectories. Specifically, this utilises the change in
state of detections from the previous to current timestep as a function
to project future positions, represented by the following state-space
equation:

\begin{equation}x_{t} = Fx_{t - 1} + w_{t}\end{equation}

Where \(x_{t}\) is the location, aspect ratio, height, and the
respective velocities of each for the track in the current frame,
\(x_{t - 1}\) represents the same variables in the previous frame, \(F\)
is a transition matrix that maps this previous state to the present
state, and \(w_{t}\) is a measure of uncertainty that is used to update
the prediction via one of several filtering methods~\cite{Rawlings2006}.

The second modification implemented in ByteTraX is minimisation of the
IoU overlap needed to meet the track association similarity threshold,
resulting in consideration of all detections proximate to the projected
track location, rather than only those with substantial overlap. This
reflects the reality that motion model predictions are frequently
limited in accuracy, and thus matching should aim to approximate the
general area distribution of a predicated track, rather than requiring
precise overlap. Specifically, the IoU between motion model predictions
and candidate detections is calculated as:

\begin{equation}IoU(D_{v},T) = \frac{\mid D_{v} \cup T \mid}{\mid D_{v} \cap T \mid}\end{equation}

In which \(D_{v} \cup T\) is the area of overlap between the predicted
track location and detection, and \(D_{v} \cap T\) is their combined
area when considered as a single shape.

The quantification of IoU-based similarity is then determined via the
following equation:

\begin{equation}C = 1 - IoU(T,D_{v})\end{equation}

Where \(C\) is a cost function, calculated as the IoU of predicted track
location \(T\) and valid detection \(D_{v}\) subtracted from 1.
Consequently, by minimising the deviation from 1 required to satisfy the
track association similarity threshold, the more robust predictions
become to uncertainty, and the fewer track breaks and ID reassignations
occur across trajectories.

Taken together, these modifications result in substantially more lenient
trajectory propagation thresholds, thus lending increased weight to
existing tracks. As such, there is a concomitant reduction in the
establishment of new track IDs, as extant tracks must drop below a
comparatively minimal association threshold before being lost. Broadly,
this aims to address the issues of erroneous track loss and ID
reassignation, while simultaneously increasing inference speed through a
simplified association algorithm.

Beyond threshold modifications, ByteTraX introduces further improvements
specific to closed systems---those in which the total number of tracks
remain consistent as objects do not enter or leave the frame. These
constitute a special case in which global information as to the existing
and expected track states can be leveraged to enhance association
decisions. To this end, ByteTraX incorporates track reconnection and
merging functions that further suppress ID reassignation, while enabling
lost track recovery beyond that achievable by the motion model alone.

\begin{table*}[htbp]
{\legendjustify\textbf{Table 1.} Comparison of attributes across benchmark datasets.
Values represent means averaged across all video sequences within each
benchmark. Metrics are selected to outline target size, density, speed,
propensity for sudden changes in acceleration and direction, occlusion
rate, divergence from linear motion model predictions, and simultaneous
class cooccurrence.\par}

\vspace{10pt}
\small
\iflatexml
\begin{tabular*}{\linewidth}{@{\extracolsep{\fill}}
  >{\centering\arraybackslash}p{0.1605\linewidth}
  >{\centering\arraybackslash}p{0.0698\linewidth}
  >{\centering\arraybackslash}p{0.0602\linewidth}
  >{\centering\arraybackslash}p{0.0965\linewidth}
  >{\centering\arraybackslash}p{0.0904\linewidth}
  >{\centering\arraybackslash}p{0.1355\linewidth}
  >{\centering\arraybackslash}p{0.0920\linewidth}
  >{\centering\arraybackslash}p{0.1287\linewidth}
  >{\centering\arraybackslash}p{0.1031\linewidth}
  >{\centering\arraybackslash}p{0.0633\linewidth}@{}}
\else
\begin{tabular}{@{}
  >{\centering\arraybackslash}p{(\linewidth - 18\tabcolsep) * \real{0.1605}}
  >{\centering\arraybackslash}p{(\linewidth - 18\tabcolsep) * \real{0.0698}}
  >{\centering\arraybackslash}p{(\linewidth - 18\tabcolsep) * \real{0.0602}}
  >{\centering\arraybackslash}p{(\linewidth - 18\tabcolsep) * \real{0.0965}}
  >{\centering\arraybackslash}p{(\linewidth - 18\tabcolsep) * \real{0.0904}}
  >{\centering\arraybackslash}p{(\linewidth - 18\tabcolsep) * \real{0.1355}}
  >{\centering\arraybackslash}p{(\linewidth - 18\tabcolsep) * \real{0.0920}}
  >{\centering\arraybackslash}p{(\linewidth - 18\tabcolsep) * \real{0.1287}}
  >{\centering\arraybackslash}p{(\linewidth - 18\tabcolsep) * \real{0.1031}}
  >{\centering\arraybackslash}p{(\linewidth - 18\tabcolsep) * \real{0.0633}}@{}}
\fi
\toprule\noalign{}
\begin{minipage}[b]{\linewidth}\centering
Dataset
\end{minipage} & \begin{minipage}[b]{\linewidth}\centering
Video Length
\end{minipage} & \begin{minipage}[b]{\linewidth}\centering
Track Count
\end{minipage} & \begin{minipage}[b]{\linewidth}\centering
Track Density (Tracks/ MP)
\end{minipage} & \begin{minipage}[b]{\linewidth}\centering
Track Speed (Px/ Frame)
\end{minipage} & \begin{minipage}[b]{\linewidth}\centering
Track Angular Acceleration (°/ Frame)
\end{minipage} & \begin{minipage}[b]{\linewidth}\centering
Track Occlusion Rate
\end{minipage} & \begin{minipage}[b]{\linewidth}\centering
Track Normalised Innovation Residual
\end{minipage} & \begin{minipage}[b]{\linewidth}\centering
Target Size (Px)
\end{minipage} & \begin{minipage}[b]{\linewidth}\centering
Class Count
\end{minipage} \\
\midrule\noalign{}
DeepSea-MOT~\cite{Barnard2025} & 599 & 35 & 7.183 & 4.170 & 36.170 & 0.268 & 0.320 &
51542.470 & 37 \\
\cmidrule{1-10}
DAMUNT~\cite{Abeysinghe2023} & 573 & 23 & 40.453 & 1.810 & 59.360 & 0.677 & 0.160 & 2019.370 &
1 \\
\cmidrule{1-10}\noalign{}
GMOT-40~\cite{Bai2021} & 237 & 49 & 14.830 & 10.040 & 21.180 & 0.561 & 0.740 &
12989.050 & 1 \\
\cmidrule{1-10}
LC-MOT~\cite{Yuan2026} & 1124 & 15 & 11.398 & 1.030 & 55.200 & 0.400 & 0.380 & 4634.940
& 1 \\
\cmidrule{1-10}
SportsMOT~\cite{Cui2023} & 575 & 14 & 12.004 & 6.070 & 27.000 & 0.593 & 0.390 &
6756.340 & 1 \\
\cmidrule{1-10}
TeamTrack~\cite{Scott2024} & 900 & 23 & 2.773 & 1.880 & 20.540 & 0.126 & 0.160 & 922.680
& 1 \\
\bottomrule\noalign{}
\iflatexml\end{tabular*}\else\end{tabular}\fi

\vspace{0.3em}
{\footnotesize Normalised innovation residuals are calculated using the standard
ByteTrack motion model, sampled from 10\% of tracks per video.}
\settabcurrentlabel
\label{table:1}
\end{table*}

\subsection{Track Reconnection}

To enable extended track reconnection, a simple search heuristic is
employed when a single track ID is lost and cannot be recovered by the
motion model. This is possible in closed systems, as the fixed maximum
track count ensures that when only a single ID is lost, any new
trajectory must be associated with the same ID. Specifically, this
heuristic associates any same-class detection within a fixed relative
distance and time threshold with the lost tracklet, regardless of motion
projections. This process is described by the following formula:

\begin{equation}\begin{array}{r}
R = \lbrack\Delta t \leq \tau \land \Delta P \leq \alpha\rbrack
\end{array}\end{equation}

Where \(R\) denotes the decision to reconnect a lost track with a new
trajectory, dependent upon their relative pairwise difference \(\Delta\)
in occurrence time \(t\), and proximity \(P\). To satisfy the
reconnection criteria, both values must fall within their respective
thresholds \(\tau\) and \(\alpha\).

In the case of multiple candidate detections, matching is filtered based
on proximity to the last known track location, with the process
repeating until a tracking lock is achieved, or the thresholds are
exhausted. To ensure robustness, this heuristic is activated only when
the standard association rules do not yield a suitable reconnection. The
advantage of this approach is that it robustly handles cases where
extreme non-linear movements lead to failure of the motion model, or
those where extended occlusion or detection failure disrupt tracklet
continuity. Additionally, this algorithm requires no knowledge of the
total track count, only whether there has been a single or multiple
losses in relation to the previous system state.

\subsection{Tracklet Merging}

To effectively resolve the issue of erroneous ID reassignation, a
merging step is implemented for screening all new tracklets. This relies
upon the logic that for a system with a fixed track count for which all
tracks are currently present, additional tracklets that overlap with
existing trajectories likely represents repeat detections of the same
objects. Consequently, new tracks that meet these criteria are
immediately merged into the existing same-class tracks with which they
overlap, thus preventing the assignation of a new ID to an established
track, even if the former exhibits higher detection confidence. This
process is governed by the following equation:

\begin{equation}\begin{array}{r}
\text{M = }IoU(T_{\text{existing}},D_{\text{new}}) \geq \tau_{\text{merge}}
\end{array}\end{equation}

In which \(\text{M}\) represents the decision to merge a new detection
\(D_{\text{new}}\) into an existing track \(T_{\text{existing}}\),
dependent upon their combined IoU meeting or exceeding the threshold
\(\tau_{\text{merge}}\).

As with the reconnection function, this heuristic activates only once
the standard association and matching steps have been completed, thus
ensuring that lost tracks are not erroneously merged into existing
trajectories. Additionally, this provides robustness to cases where
tracks overlap or otherwise closely intersect, as the merging function
only applies to those detections that cannot be reliably assigned via
the motion model or reconnection steps.

To facilitate adaptable usage, the aforementioned heuristics can be
activated in addition to the standard architecture when tracking within
closed systems. This is important for ensuring flexibility, as they may
be unsuitable for use in open systems where information on track counts
and states is incomplete. Additionally, the ByteTraX algorithm aims to
deliver performance improvements in closed systems even without these
functions, meaning that all aspects can be implemented optionally or
integrated into additional frameworks.

\section{Experiments}

\begin{figure*}[t]
{\centering\includegraphics[width=\textwidth]{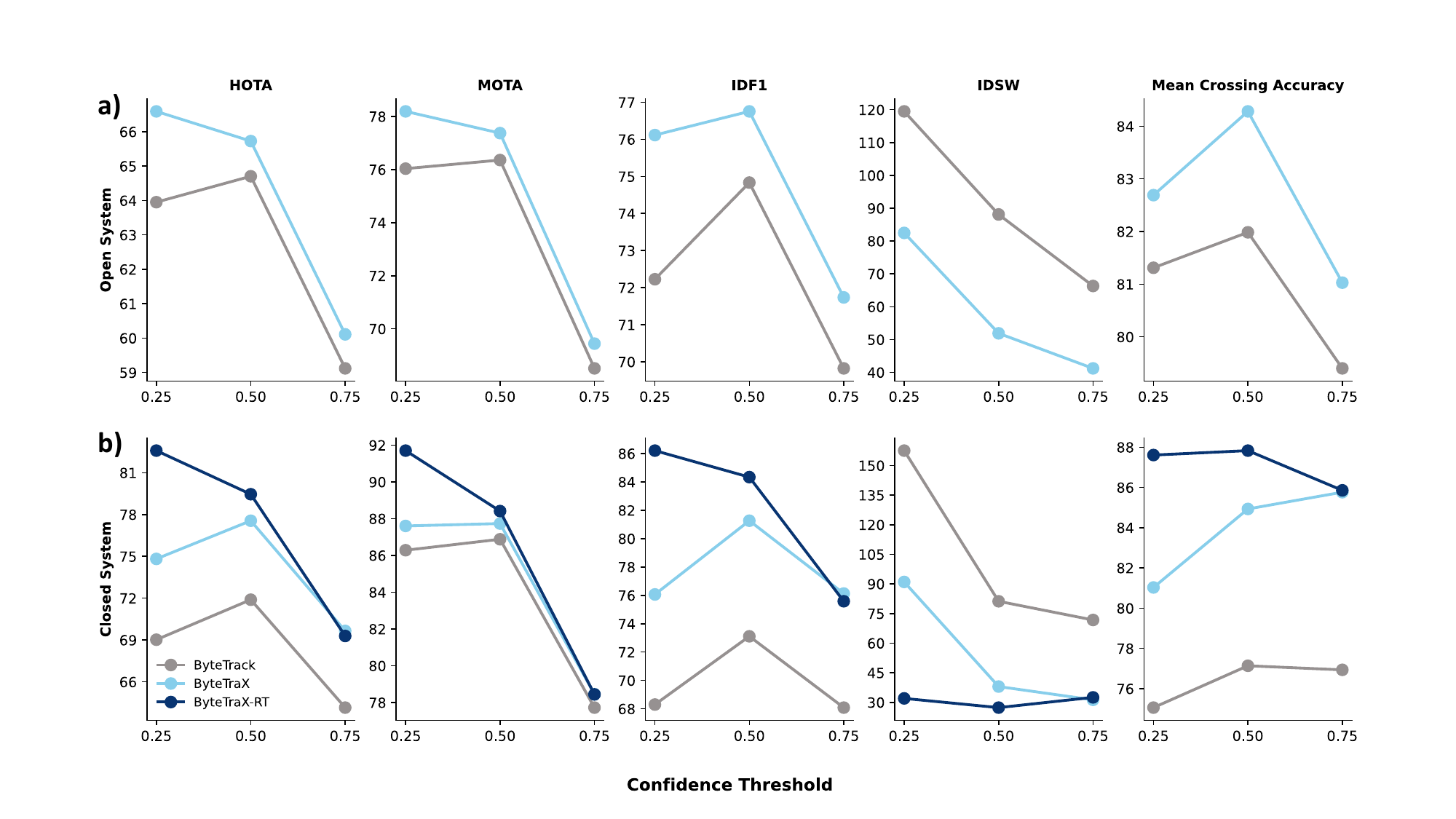}\par}

\vspace{10pt}
{\legendjustify\textbf{Figure 3.} (a) Comparison of tracking performance across
detection confidence thresholds for open system benchmarks
(\emph{N}=53). Each point represents the combined mean score for
GMOT-40, SportsMOT, and DeepSea-MOT, coloured by tracker (ByteTrack,
grey; ByteTraX, light blue). (b) Comparison of tracking performance
across detection confidence thresholds for closed system benchmarks
(\emph{N}=15). Each point represents the mean score averaged across
LC-MOT, TeamTrack, and DAMUNT, coloured by tracker (ByteTrack, grey;
ByteTraX, light blue; ByteTraX-RT, dark blue).\par}
\setfigcurrentlabel
\label{fig:3}
\end{figure*}

\begin{table*}[t]
{\legendjustify\textbf{Table 2.} Comparison of tracking performance across open and
closed systems. Values represent combined means averaged over all
benchmark datasets and detection confidence thresholds belonging to each
system type, with the highest scoring results shown in bold. The overall
tracking algorithm rankings are determined based upon the proportion of
top scoring metrics achieved by each.\par}

\vspace{10pt}
\small
\iflatexml
\begin{tabular*}{\linewidth}{@{\extracolsep{\fill}}
  >{\centering\arraybackslash}p{0.1431\linewidth}
  >{\centering\arraybackslash}p{0.1756\linewidth}
  >{\centering\arraybackslash}p{0.1043\linewidth}
  >{\centering\arraybackslash}p{0.0971\linewidth}
  >{\centering\arraybackslash}p{0.1240\linewidth}
  >{\centering\arraybackslash}p{0.1702\linewidth}
  >{\centering\arraybackslash}p{0.0892\linewidth}
  >{\centering\arraybackslash}p{0.0966\linewidth}@{}}
\else
\begin{tabular}{@{}
  >{\centering\arraybackslash}p{(\linewidth - 14\tabcolsep) * \real{0.1431}}
  >{\centering\arraybackslash}p{(\linewidth - 14\tabcolsep) * \real{0.1756}}
  >{\centering\arraybackslash}p{(\linewidth - 14\tabcolsep) * \real{0.1043}}
  >{\centering\arraybackslash}p{(\linewidth - 14\tabcolsep) * \real{0.0971}}
  >{\centering\arraybackslash}p{(\linewidth - 14\tabcolsep) * \real{0.1240}}
  >{\centering\arraybackslash}p{(\linewidth - 14\tabcolsep) * \real{0.1702}}
  >{\centering\arraybackslash}p{(\linewidth - 14\tabcolsep) * \real{0.0892}}
  >{\centering\arraybackslash}p{(\linewidth - 14\tabcolsep) * \real{0.0966}}@{}}
\fi
\toprule\noalign{}
\begin{minipage}[b]{\linewidth}\centering
System Type
\end{minipage} & \begin{minipage}[b]{\linewidth}\centering
Tracker
\end{minipage} & \begin{minipage}[b]{\linewidth}\centering
HOTA
\end{minipage} & \begin{minipage}[b]{\linewidth}\centering
MOTA
\end{minipage} & \begin{minipage}[b]{\linewidth}\centering
IDF1
\end{minipage} & \begin{minipage}[b]{\linewidth}\centering
Crossing Accuracy
\end{minipage} & \begin{minipage}[b]{\linewidth}\centering
IDSW
\end{minipage} & \begin{minipage}[b]{\linewidth}\centering
FPS
\end{minipage} \\
\midrule\noalign{}
Open & ByteTrack & 62.593 & 73.634 & 72.294 & 80.899 & 91 & 44.328 \\
& \textbf{ByteTraX} & \textbf{64.143} & \textbf{75.002} &
\textbf{74.869} & \textbf{82.667} & \textbf{58} & \textbf{49.498} \\
\cmidrule{1-8}\noalign{}
Closed & ByteTrack & 68.356 & 83.629 & 69.832 & 76.379 & 103 & 51.601 \\
& ByteTraX & 74.009 & 84.597 & 77.819 & 83.909 & 53 & 58.645 \\
& \textbf{ByteTraX-RT} & \textbf{77.111} & \textbf{86.185} &
\textbf{82.043} & \textbf{87.098} & \textbf{30} & \textbf{59.265} \\
\bottomrule\noalign{}
\iflatexml\end{tabular*}\else\end{tabular}\fi
\settabcurrentlabel
\label{table:2}
\end{table*}

\subsection{Datasets}

I utilise six standardised benchmarks to evaluate the performance of
ByteTraX in comparison to that of the ByteTrack baseline. These encompass a diversity of
tracking scenarios, target types, and motion patterns, while providing
detailed ground truth annotations for both bounding boxes and track
identities. The first three of these---GMOT-40~\cite{Bai2021}, SportsMOT~\cite{Cui2023}, and
DeepSea-MOT~\cite{Barnard2025}---represent open systems in which tracks can enter and leave
the frame (Fig.~\hyperref[fig:2]{2a}); while the latter three---LC-MOT~\cite{Yuan2026}, TeamTrack~\cite{Scott2024}, and
DAMUNT~\cite{Abeysinghe2023}---are closed systems in which all tracks remain in view across
videos (Fig.~\hyperref[fig:2]{2b}). While the vast scope of potential real-world
deployments cannot feasibly be encompassed by benchmarking
alone~\cite{OSheaWheller2026}, the aforementioned datasets allow for
relative performance comparisons across widely divergent tracking
scenarios. Illustrative of this, sequences contain from 4 to 100
simultaneously cooccurring tracks, encompass mean trajectory speeds
between 0.2 and 89 pixels per second, and feature occlusion rates
ranging from 0.010 to 0.983 (Table~\hyperref[table:1]{1}).

Such variance is further supported by a diversity of tracking target
appearances, outlined as follows. GMOT-40 encompasses scenes with
targets including humans, aircraft, cars, fish, insects, livestock, and
boats, with the aim of testing generic multi-category tracking
performance~\cite{Bai2021}. TeamTrack and SportsMOT focus on
tracking Football, Basketball, and Volleyball players, using footage
filmed either laterally with high camera motion~\cite{Cui2023}, or
directly from above~\cite{Scott2024}. DAMUNT utilises colonies of the
weaver ant \emph{Oecophylla smaragdina}, and carpenter ant
\emph{Camponotus aeneopilosus} in the lab~\cite{Abeysinghe2023}, combining
high target density with rapid non-linear movement patterns. DeepSea-MOT
encompasses ROV footage from the deep ocean, providing an extensive
range of \textgreater40 co-occurring species classes from varied taxa
across midwater and benthic environments~\cite{Barnard2025}. Finally,
LC-MOT consists of videos of the fish \emph{Larimichthys crocea} filmed
from above in rearing tanks, yielding dense aggregations of
visually-similar individuals with frequent occlusions and severe surface
reflectance~\cite{Yuan2026}.

\subsection{Metrics}

To comprehensively evaluate tracking performance, I employ
HOTA~\cite{Luiten2020}, MOTA~\cite{Bernardin2008},
IDF1~\cite{Ristani2016}, IDSW~\cite{Dendorfer2020}, excess ID
count~\cite{Ihara2025}, and FPS~\cite{Dendorfer2020}. These metrics
encompass tracking accuracy, identity preservation, and computational
efficiency, enabling assessment of ByteTraX's potential to reduce
erroneous ID reassignation, along with any resultant impacts on overall
performance. Specifically, HOTA is selected to quantify combined
tracking accuracy as it balances detection, association, and
localisation performance~\cite{Luiten2020}, while MOTA provides a
more detection-centric measure, and is included due to its extensive
usage in earlier work~\cite{Bernardin2008}. Identity preservation and
track switching are assessed via IDF1, IDSW, and excess ID count, as
these allow for evaluation of the proportion of detections correctly
assigned to their corresponding ground truth IDs~\cite{Ristani2016},
the absolute number of ID switches and supersedure
events~\cite{Bernardin2008}, and the resultant accumulation of erroneous
IDs in comparison to ground truth counts~\cite{Stanojevic2025,Zhang2024}.
Finally, FPS is utilised as a general proxy for computational
efficiency, as when hardware and detection model parameters are
standardised, differences in frame rate reliably reflect variance in
algorithmic processing speed~\cite{Wang2020}.

In addition to the aforementioned multi-object tracking metrics, I
integrate a simple line crossing accuracy measure to validate
performance in a common counting task. This utilises six evenly spaced
vertical and horizontal lines overlaid across the video frame, enabling
quantification of trajectory crossing counts in comparison to ground
truth values (Fig.~\hyperref[fig:2]{2}). Such a measure encapsulates the ubiquitous use
case of counting-by-tracking~\cite{Makar2025,Bisio2022,Sattarzadeh2025}, ensuring direct
applicability to real-world deployments. In doing so, this links
fundamental tracking metrics with higher-level counting outcomes, while
furnishing an evaluation pipeline for use in future work. Together,
these measures enable assessment of tracking performance from a fine to
global scale, encompassing both standard computer vision metrics and a
composite downstream task.

\subsection{Implementation}

Trials employed YOLO26n~\cite{Jocher2026} as the common detection
architecture, with a separate model trained on each benchmark for 100
epochs, thus serving as a consistent detection front-end for use across
both tracking algorithms. As training and testing portions differed
between datasets, a standard 40:10:50 training, validation, and testing
split was adopted in all cases. This ensured that models were
well-optimised to their respective testing benchmarks, allowing
assessments to focus on identity association and tracking performance
independently of detector-based limitations. While this necessarily
limits comparisons to the two tracking algorithms described here, it
ensures reliable and robust performance evaluations that are broadly
generalisable across a range of use cases.

For the open system datasets, ByteTraX was evaluated only in its
standard configuration, while in the closed system datasets, it was
additionally tested with the track reconnection and merging features
enabled---the latter henceforth being referred to as `ByteTraX-RT'. This
allowed for additive evaluation of both the core ByteTraX architecture
and specific closed system enhancements in comparison to the ByteTrack
baseline. Metric extraction utilised the established `TrackEval'
framework~\cite{Dendorfer2020,Luiten2020a,Aouini2024}, yielding HOTA~\cite{Luiten2020},
CLEAR~\cite{Bernardin2008}, Identity~\cite{Ristani2016}, and
VACE~\cite{Manohar2006} metrics, along with a bespoke analysis pipeline
for recording line crossing accuracy measures. To explore performance
stability across input values, separate runs were undertaken at
detection confidence thresholds of 0.25, 0.50, and 0.75 for each
permutation of tracker and video sequence. All other model parameters
remained identical across testing regimes to ensure comparability, with
an IoU threshold of 0.70, image downsampling resolution of 640x640, and
inference arguments set to the defaults present in
YOLO26~\cite{Jocher2026}. Training and testing was conducted
using a NVIDIA Tesla A100 Tensor Core GPU (NVIDIA) in a high-RAM runtime
configuration, with associated FPS values derived from the same
hardware. All statistical analyses were performed in Python (release v.
3.9.12).

\subsection{Comparison with ByteTrack}

As the principle aim of ByteTraX is not to challenge the
state-of-the-art, but to improve upon the widely-established ByteTrack
baseline, I focus my comparisons on these two architectures exclusively. Results
are broadly divided into those from open and closed system benchmarks,
and all assessments utilise the same upstream YOLO26 models across
tracking algorithms to ensure parity.

\begin{figure*}[t]
{\centering\includegraphics[width=\textwidth]{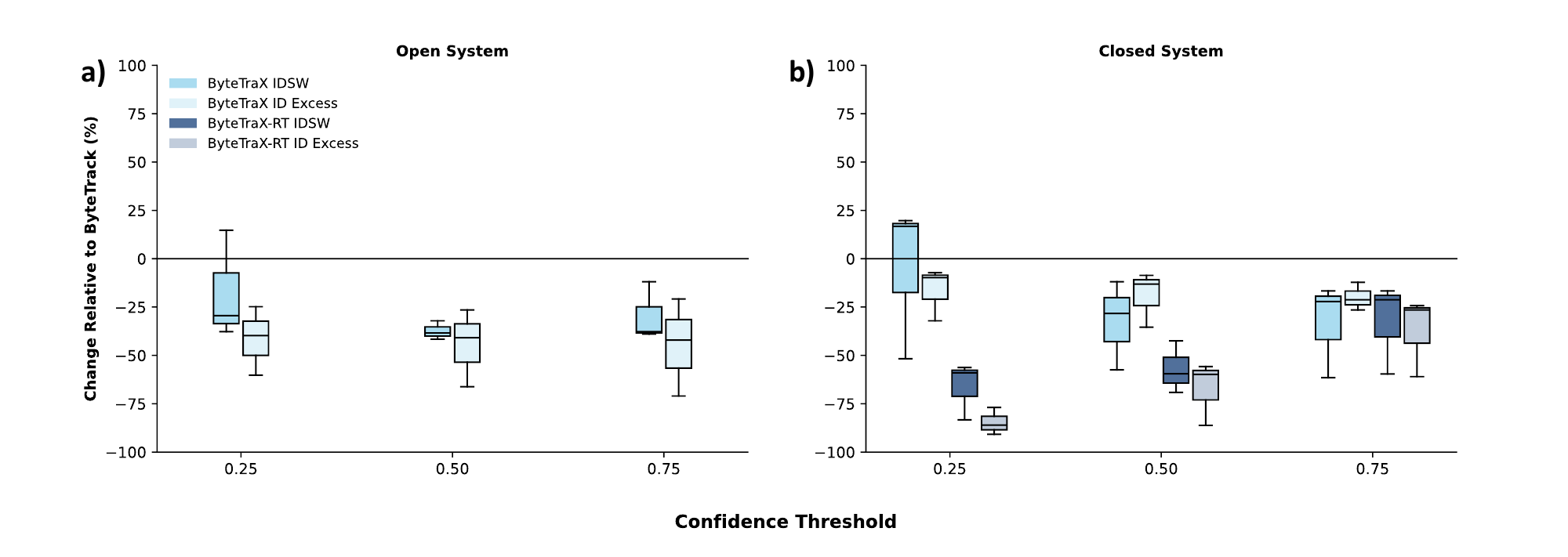}\par}

\vspace{10pt}
{\legendjustify\textbf{Figure 4.} (a) Change in IDSW and excess ID counts relative to
ByteTrack across detection confidence thresholds for open system
benchmarks (\emph{N}=53). Boxplots indicate the median percentage change
in proportion to the corresponding value for ByteTrack, averaged across
GMOT-40, SportsMOT, and DeepSea-MOT, and coloured by tracker (ByteTraX, light blue) and
metric (IDSW, opaque; ID Excess, transparent). (b) Change in IDSW and
excess ID counts relative to ByteTrack across detection confidence
thresholds for closed system benchmarks (\emph{N}=15). Boxplots again
indicate the median percentage change in proportion to the corresponding
value for ByteTrack, averaged across LC-MOT, TeamTrack, and DAMUNT, and coloured by
tracker (ByteTraX, light blue; ByteTraX-RT, dark blue) and metric (IDSW,
opaque; ID Excess, transparent). Error bars represent 1.5 times the
interquartile range from the median. Positive values indicate an increase
in the given metric, negative values a decrease, and a score of 0 absolute
parity with ByteTrack. ID excess is defined as the number of additional
IDs generated by a tracker in comparison to ground truth, while IDSW is
a count of all instances in which the same ground truth tracks are
assigned sequentially different IDs.\par}
\setfigcurrentlabel
\label{fig:4}
\end{figure*}

\mysection{Open System Benchmarks}
When considering open system benchmarks, ByteTraX substantially
outperforms ByteTrack in overall tracking accuracy and speed, achieving
a mean increase in HOTA of 1.550, MOTA of 1.368, and FPS of 5.170 (Fig.~\hyperref[fig:3]{3a}, Table~\hyperref[table:2]{2}). This is driven by markedly enhanced identity preservation,
reflected in an IDF1 score increase of 2.575, and crucially, a
\textgreater36\% reduction in identity switches, with mean IDSW dropping
from 91 to 58, and excess IDs from 1013 to 447 (Fig.~\hyperref[fig:3]{3a}, Fig.~\hyperref[fig:4]{4}, Table~\hyperref[table:2]{2}). Performance improvements are further reflected in elevated line
crossing accuracy, with an aggregate increase in this metric of 1.768\%
over the ByteTrack baseline (Fig.~\hyperref[fig:3]{3a}, Table~\hyperref[table:2]{2}).

These trends are generally consistent across confidence thresholds,
although with a slight increase in relative performance gains at a
threshold of 0.25, potentially indicating the increased cost of
uncertainty for ByteTrack when presented with additional low confidence
detections (Fig.~\hyperref[fig:3]{3a}). At the dataset level, results for both GMOT-40 and
SportsMOT are broadly comparable, with ByteTraX's ability to reduce
erroneous identity reassignments and switches driving similar
performance improvements in HOTA, MOTA, IDF1, IDSW, and excess ID counts
(Table~\hyperref[table:2]{2}, Fig.~\hyperref[fig:4]{4a}). Notably however, despite the same being true for
DeepSea-MOT in terms of HOTA, IDF1, and crossing accuracy scores,
ByteTraX did not achieve meaningfully elevated MOTA or reduced IDSW
rates (Table~\hyperref[table:3]{3}). Investigation across confidence levels revealed this to
be primarily driven by deficits at a threshold of 0.25, with IDSW rates
dropping only at thresholds of 0.50 and 0.75 (Fig.~\hyperref[fig:4]{4a}).

\begin{table*}[t]
{\legendjustify\textbf{Table 3.} Comparison of tracking performance across benchmark
datasets. Values represent combined means averaged over all detection
confidence thresholds, with the highest scoring results for each dataset
shown in bold. The overall tracking algorithm rankings are determined
based upon the proportion of top scoring metrics belonging to each.\par}

\vspace{10pt}
\small
\iflatexml
\begin{tabular*}{\linewidth}{@{\extracolsep{\fill}}
  >{\centering\arraybackslash}p{0.0740\linewidth}
  >{\centering\arraybackslash}p{0.2000\linewidth}
  >{\centering\arraybackslash}p{0.1950\linewidth}
  >{\centering\arraybackslash}p{0.0924\linewidth}
  >{\centering\arraybackslash}p{0.0913\linewidth}
  >{\centering\arraybackslash}p{0.0879\linewidth}
  >{\centering\arraybackslash}p{0.1144\linewidth}
  >{\centering\arraybackslash}p{0.0541\linewidth}
  >{\centering\arraybackslash}p{0.0908\linewidth}@{}}
\else
\begin{tabular}{@{}
  >{\centering\arraybackslash}p{(\linewidth - 16\tabcolsep) * \real{0.0740}}
  >{\centering\arraybackslash}p{(\linewidth - 16\tabcolsep) * \real{0.2000}}
  >{\centering\arraybackslash}p{(\linewidth - 16\tabcolsep) * \real{0.1950}}
  >{\centering\arraybackslash}p{(\linewidth - 16\tabcolsep) * \real{0.0924}}
  >{\centering\arraybackslash}p{(\linewidth - 16\tabcolsep) * \real{0.0913}}
  >{\centering\arraybackslash}p{(\linewidth - 16\tabcolsep) * \real{0.0879}}
  >{\centering\arraybackslash}p{(\linewidth - 16\tabcolsep) * \real{0.1144}}
  >{\centering\arraybackslash}p{(\linewidth - 16\tabcolsep) * \real{0.0541}}
  >{\centering\arraybackslash}p{(\linewidth - 16\tabcolsep) * \real{0.0908}}@{}}
\fi
\toprule\noalign{}
\begin{minipage}[b]{\linewidth}\centering
System Type
\end{minipage} & \begin{minipage}[b]{\linewidth}\centering
Dataset
\end{minipage} & \begin{minipage}[b]{\linewidth}\centering
Tracker
\end{minipage} & \begin{minipage}[b]{\linewidth}\centering
HOTA
\end{minipage} & \begin{minipage}[b]{\linewidth}\centering
MOTA
\end{minipage} & \begin{minipage}[b]{\linewidth}\centering
IDF1
\end{minipage} & \begin{minipage}[b]{\linewidth}\centering
Crossing Accuracy
\end{minipage} & \begin{minipage}[b]{\linewidth}\centering
IDSW
\end{minipage} & \begin{minipage}[b]{\linewidth}\centering
FPS
\end{minipage} \\
\midrule\noalign{}
Open & GMOT-40~\cite{Bai2021} & ByteTrack & 62.557 & 77.971 & 74.519 & 82.880 & 107 &
46.476 \\
& & \textbf{ByteTraX} & \textbf{64.379} & \textbf{79.775} &
\textbf{77.562} & \textbf{84.577} & \textbf{65} & \textbf{51.720} \\
\cmidrule{2-9}\noalign{}
& SportsMOT~\cite{Cui2023} & ByteTrack & 62.049 & 90.146 & 71.400 & 78.559 & 67 &
57.249 \\
& & \textbf{ByteTraX} & \textbf{63.304} & \textbf{90.930} &
\textbf{74.279} & \textbf{82.086} & \textbf{44} & \textbf{66.475} \\
\cmidrule{2-9}\noalign{}
& DeepSea-MOT~\cite{Barnard2025} & ByteTrack & 62.898 & \textbf{51.331} & 63.661 & 73.679 &
\textbf{37} & 31.857 \\
& & \textbf{ByteTraX} & \textbf{63.452} & 51.128 & \textbf{64.272} &
\textbf{75.203} & 38 & \textbf{35.517} \\
\cmidrule{1-9}\noalign{}
Closed & DAMUNT~\cite{Abeysinghe2023} & ByteTrack & 68.866 & 85.589 & 65.165 & 77.575 & 150 &
53.799 \\
& & ByteTraX & 77.856 & 87.25 & 77.366 & 89.004 & 67 & 61.621 \\
& & \textbf{ByteTraX-RT} & \textbf{81.962} & \textbf{89.242} &
\textbf{82.651} & \textbf{94.048} & \textbf{39} & \textbf{62.845} \\
\cmidrule{2-9}\noalign{}
& LC-MOT~\cite{Yuan2026} & ByteTrack & 79.355 & 92.627 & 89.021 & 68.671 & 11 &
54.558 \\
& & ByteTraX & 79.574 & 92.587 & 89.677 & \textbf{69.994} & 11 &
61.577 \\
& & \textbf{ByteTraX-RT} & \textbf{81.081} & \textbf{93.391} &
\textbf{92.146} & 69.788 & \textbf{5} & \textbf{60.512} \\
\cmidrule{2-9}\noalign{}
& TeamTrack~\cite{Scott2024} & ByteTrack & 44.062 & 56.817 & 52.456 & 86.409 & 82 &
35.797 \\
& & ByteTraX & 45.572 & 56.677 & 56.137 & 88.808 & 79 & 39.383 \\
& & \textbf{ByteTraX-RT} & \textbf{47.344} & \textbf{58.017} &
\textbf{59.104} & \textbf{90.444} & \textbf{45} & \textbf{40.656} \\
\bottomrule\noalign{}
\iflatexml\end{tabular*}\else\end{tabular}\fi
\settabcurrentlabel
\label{table:3}
\end{table*}

\mysection{Closed System Benchmarks}
In closed systems, ByteTraX again demonstrates superior HOTA, MOTA,
IDF1, and FPS in comparison to ByteTrack, along with reduced IDSW scores
and excess ID counts (Fig.~\hyperref[fig:3]{3b} and \hyperref[fig:4]{4b}, Table~\hyperref[table:2]{2}). Specifically, HOTA and
IDF1 scores are increased by 5.653 and 7.987 respectively, while MOTA is
elevated by 0.968, suggesting that performance gains are primarily due
to improvements in association rather than detection accuracy. This is
further supported by a \textgreater48\% reduction in identity
switches---with mean IDSW score dropping from 103 to 53---and a decline
in excess IDs from 137 to 109 (Fig.~\hyperref[fig:3]{3b} and \hyperref[fig:4]{4b}, Table~\hyperref[table:2]{2}). In combination,
these improvements yield a mean line crossing accuracy increase of
7.530\%, while elevating FPS by 7.044 (Fig.~\hyperref[fig:3]{3b}, Table~\hyperref[table:2]{2}).

Activation of the track reconnection and merging functions improves
performance further, with ByteTraX-RT yielding superior scores across
all metrics. When compared to ByteTrack, this manifests as an increase
in HOTA of 8.755, MOTA of 2.556, and FPS of 7.664, while improvements
over the standard ByteTraX configuration are reflected in an increase in
HOTA of 3.102, MOTA of 1.588, and FPS of 0.620 (Fig.~\hyperref[fig:3]{3b}, Table~\hyperref[table:2]{2}).
Identity preservation metrics are likewise enhanced, with IDF1 score
increases of 12.211 and 4.224 in comparison to ByteTrack and ByteTraX
respectively, and a drop in mean IDSW from 103 and 53 to 30 (Fig.~\hyperref[fig:3]{3b},
Table~\hyperref[table:2]{2}). Excess ID counts yield a similar pattern, falling to
36---equating to \textgreater73\% and \textgreater66\% reductions over
ByteTrack and ByteTraX (Fig.~\hyperref[fig:4]{4b}). Notably, such enhanced performance is
reflected in mean line crossing accuracy increases of 10.719\% over
ByteTrack, and 3.189\% over the standard ByteTraX implementation (Fig.~\hyperref[fig:3]{3b}, Table~\hyperref[table:2]{2}).
Comparisons across confidence thresholds reveal a notable trend, this
being that the performance benefits of ByteTraX-RT over ByteTraX are
comparatively diminished at a threshold of 0.75 (Fig.~\hyperref[fig:3]{3b} and \hyperref[fig:4]{4b}). This
may be due to refinements in the detections generated by the models
themselves, as a greater proportion of false positives are filtered out,
thus reducing the need for tracklet merging. Further, when considering
dataset-level metrics, ByteTraX and ByteTraX-RT deliver the greatest
performance improvements for DAMUNT and TeamTrack, while those for
LC-MOT are comparatively modest (Table~\hyperref[table:3]{3}). Although it is not possible
to attribute this to a single factor, the former two benchmarks do
exhibit higher excess ID counts and lower HOTA scores for ByteTrack
paired with reduced motion model innovation residuals, indicating a
greater potential for performance gains when track breaks and erroneous
ID reassignments are curtailed (Table~\hyperref[table:1]{1}, Table~\hyperref[table:3]{3}).

\section{Conclusion}

My results demonstrate that ByteTraX delivers substantially improved
performance and speed across a range of diverse tracking
scenarios---from trajectory resolution in insect colonies,
to species quantification in the deep ocean. This is achieved via
enhanced association and matching thresholds, combined with merging and
reconnection functions that limit the impact of detection confidence
upon track continuity. It should be noted that ByteTraX does not aim to
challenge the state-of-the-art in multi-object tracking, but rather to
improve upon the current most widely-utilised architecture---this being
ByteTrack. As such, I make available the full ByteTraX integration pipeline
for deployment with YOLO26, along with the source code required for
expansion into additional detection frameworks. In sum,
by combining enhanced accuracy and speed with ease-of-integration,
ByteTraX aims to advance the current baseline for real-time tracking at
scale, thus delivering fundamentally improved performance across a broad
range of applications.

\section{Acknowledgements}

I am grateful to Simon R. Nilsson and Katie I. Murray for their useful
comments regarding the manuscript.

\section{Code availability}

All supporting data, source code, integration pipelines, evaluation
scripts, and deployment instructions are available under an MIT license
at:
\url{https://github.com/Toshea111/ByteTraX}.

\iflatexml
\bibliographystyle{unsrtnat}
\makeatletter
\@ifundefined{lx@bibliography}{\bibliography{refs}}{\lx@bibliography{refs}}
\makeatother
\else
\bibliographystyle{ieeenat_fullname}
{\small
\renewcommand{\bibsection}{\section{References}}
\bibliography{refs}
}
\fi

\end{document}
}

\end{document}